\documentclass[10pt,twocolumn,letterpaper]{article}

\usepackage{cvpr}              

\usepackage{graphicx}
\usepackage{amsmath}
\usepackage{amssymb}
\usepackage{booktabs}

\usepackage{bm}
\usepackage{enumerate}
\usepackage{multirow}

\usepackage[pagebackref,breaklinks,colorlinks]{hyperref}

\usepackage[capitalize]{cleveref}
\crefname{section}{Sec.}{Secs.}
\Crefname{section}{Section}{Sections}
\Crefname{table}{Table}{Tables}
\crefname{table}{Tab.}{Tabs.}

\def\cvprPaperID{*****} 
\def\confName{CVPR}
\def\confYear{2022}

\begin{document}

\title{Music-driven Dance Regeneration with Controllable Key Pose Constraints}


\author{
Junfu Pu \quad\quad\quad
Ying Shan
\\
ARC Lab, Tencent PCG \\
jevinpu@tencent.com,
yingsshan@tencent.com
}
\twocolumn[{%
\renewcommand\twocolumn[1][]{#1}%
\maketitle
\begin{center}
\vspace{-0.5cm}
    \centering
    \captionsetup{type=figure}
    \includegraphics[width=\textwidth]{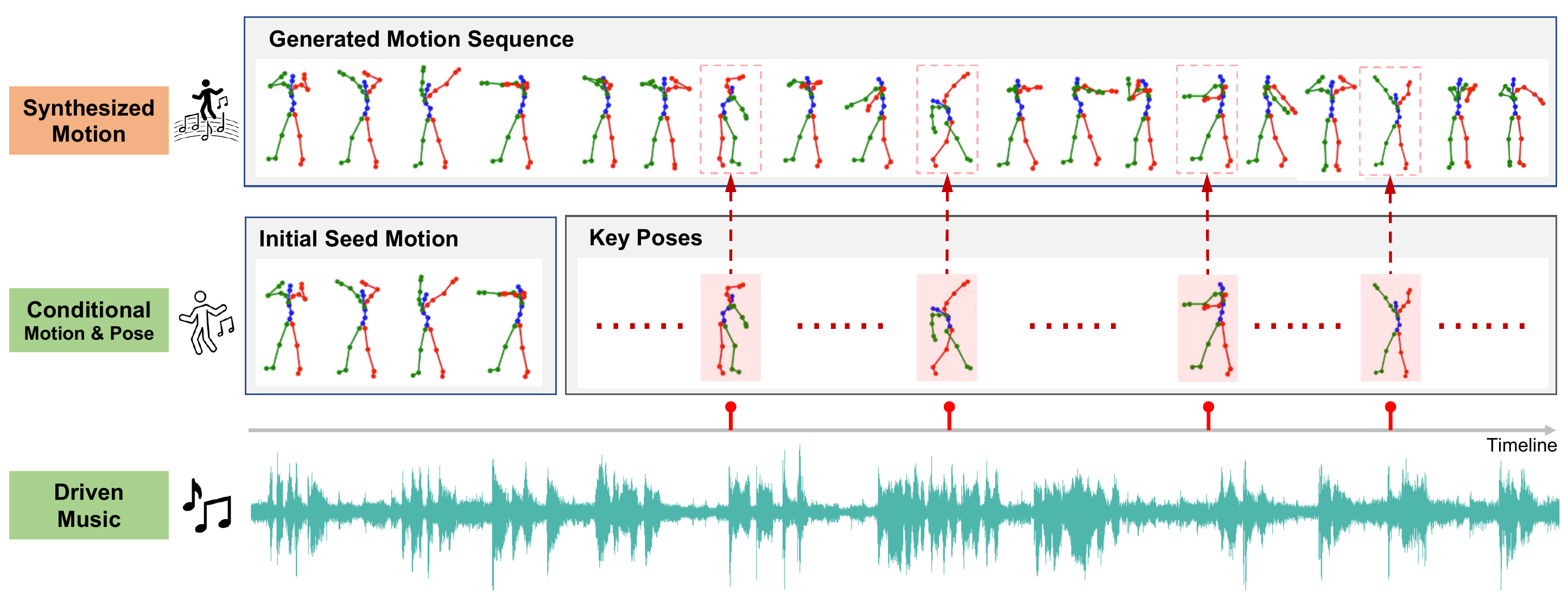}
    \captionof{figure}{Task illustration of controllable music-driven dance motion generation. Given a piece of music and a clip of initial seed motion clip, alone with some key poses which expected to be appeared in the synthesized motion sequence, our model can generate realistic and rhythmic dance motions. The output motions are controlled by the key poses at different timestamp, which means more customization available for users.}
\end{center}%
}]

\begin{abstract}
\vspace{-0.25cm}
In this paper, we propose a novel framework for music-driven dance motion synthesis with controllable key pose constraint. 
In contrast to methods that generate dance motion sequences only based on music without any other controllable conditions,
this work targets on synthesizing high-quality dance motion driven by music as well as customized poses performed by users.
Our model involves two single-modal transformer encoders for music and motion representations and a cross-modal transformer decoder for dance motions generation. 
The cross-modal transformer decoder achieves the capability of synthesizing smooth dance motion sequences, which keeps a consistency with key poses at corresponding positions, by introducing the local neighbor position embedding.
Such mechanism makes the decoder more sensitive to key poses and the corresponding positions.
Our dance synthesis model achieves satisfactory performance both on quantitative and qualitative evaluations with extensive experiments, which demonstrates the effectiveness of our proposed method.
\end{abstract}

\section{Introduction}
\label{sec:intro}
Listening to the music with inspirational touching melody, suppose that you are a performer who is always full of passion to dance rhythmically with music 
for the pleasure of experiencing the body and emotion expression.
Back to reality, however, such kind of circumstance seems to be most likely appeared in our imagination since most of us have no idea about how to dance due to the lack of professional dance training.
It would be much more encouraging if we develop a dance machine, which is able to generate realistic dance motion sequences based on music with several customized key poses provided by users.
In this work, our goal is to take a piece of music as well as some key poses and synthesize continuous human dance motion sequences.
The generated motion sequences are expected to not only follow the rhythm of music, but also keep a consistency with the key poses.
To this end, we design a novel Transformer-based architecture which involves two single-modal transformer encoders for music and initial seed motion embedding, and a cross-modal transformer decoder for motion generation controlled by key pose constraints.
We show that our model stably generate various dance sequences with different key poses driven by the same music.

Dance consists of rhythmic movements along with music, which is of aesthetic value.
Dance is a kind of art which is performed in many cultures for social, celebration, entertainment, \etc.
People express their emotions by dancing with different types of music.
Recently, with the popularization of online media platforms, more and more users try to record dancing videos performed by theirselves.
For the consideration of entertainment or business, they upload and share the dancing video on  various online social applications. 
However, dancing is a type of art that is full of skills.
A good dancer need to be professionally trained in different aspects with expensive cost, including basic dance motions as well as choreography, \etc.
Such character of dance makes it extremely difficult for most people to perform an impressive dance performance. 
On the one hand, dance motion synthesis with music have a wide range of applications including digital human, dance assistant, gaming, \etc
On the other hand, researching on dance synthesis could help researcher to explore better techniques for cross-modal sequence-to-sequence generation tasks.
Hence, the idea of developing an automatically dance motion generation system with computer vision and artificial intelligence techniques is naturally proposed for the potential commercial value as well as academic explorations.

When it comes to music-driven dance generation with computer vision and computer graphics techniques in deep learning, music-motion paired data play significant roles.
In general, a large quantity of clean data is necessary to train a deep neural network.
However, such data is collected with motion capture devices while the professional dancer is performing, which makes the process tedious and expensive.
Regenerating dance motions within limited music-motion pairs is one of the most efficient ways to enlarge the dataset with a lower cost.
Our method can regenerate a sequence of new dance motion by randomly sampling key poses at different positions.
The synthesized dance motion matches the original music while varying from pose to pose.

Synthesizing dance motions with music has been researched for many years and is becoming a hot topic in motion synthesis.
Existing methods on dance synthesis are roughly grouped into two categories, \ie, 
retrieval based frameworks with motion graph~\cite{arikan2002interactive,shiratori2006dancing,kim2003rhythmic,kim2006making,chen2021choreomaster} 
and deep generative model based methods~\cite{alemi2017groovenet,tang2018dance,li2021ai,li2021dancenet3d,huang2020dance,zhang2021dance,ren2019music,sun2020deepdance,lee2019dancing,ren2020self}.
Early works mainly focus on motion graph.
The key idea of such framework is to generate motions by finding an optimal path in a pre-built motion graph.
Each node in the graph represents a motion clip, while the directed edge corresponds to the cost between two associated motion clips, 
considering the correlation and consistency of transition.
Kim \etal~\cite{kim2003rhythmic} propose to synthesize a new motion from unlabelled example motions, which preserves the rhythmic pattern, by traversing the motion graph.
Aristidou \etal~\cite{aristidou2018style} attempt to synthesize style-coherent animation accounting for stylistic variations of the movement.
The up-to-date dance generation system with motion graph is proposed in~\cite{chen2021choreomaster}.
In this work, Chen \etal train a deep neural network to learn choreomusical embedding and incorporate such embedding into a novel choreography-oriented graph-based motion synthesis framework, with various choreographic rules considered.
Although graph based methods have explicit explanation while the optimal path in graph is straightforward, 
the drawbacks of such methods still remain.
In fact, the output dance synthesized with motion graph is in essentially a composition of existing motion clips in database, 
which imposes restrictions on the diversity of dance.

Recently, benefiting from the development of deep learning in cross-modal understanding and sequence-to-sequence generation, 
deep generative model has become a standard scheme for music-driven dance synthesis.
Tang \etal~\cite{tang2018dance} use LSTM-autoencoder to synthesize dance choreograph with temporal indexes.
To generate dance with controllable style, Zhang \etal~\cite{zhang2021dance} propose to transfer dance styles to dance motions, in terms of generating dance with specific style. 
Transformer has shown powerful capability in sequence modeling. 
Several works apply transformers into dance synthesis~\cite{huang2020dance,li2021ai,li2021dancenet3d,li2022danceformer}.
To generate long dance sequence, Huang \etal~\cite{huang2020dance} propose a novel curriculum learning strategy to alleviate error accumulation of auto-regressive models.
Aristidou \etal~\cite{aristidou2022rhythm} present a neural framework to generate dance motion driven by music that form a global structure respecting the culture of the dance genre.
Duan \etal~\cite{duanautomatic2021} collect a large scale dancing dataset with labeled and unlabeled music.
Besides, Li \etal~\cite{li2021ai} release a new public multi-modal dataset of 3D dance motion and music, which makes a large progress of dance motion synthesis research.

In this work, they also propose a full-attention cross-modal transformer network for 3D dance motion generation.
Generating dance with deep neural networks has greater diversity in comparison with graph based methods.
However, the fundamental shortcoming of such category of methods with deep neural networks is the lack of controllability.
In our proposed method, we attempt to generate dance motions with poses expected to appear in output dance, which makes the generating process explicit and controllable.

\begin{figure*}[ht!]
  \centering
  \includegraphics[width=1.0\linewidth]{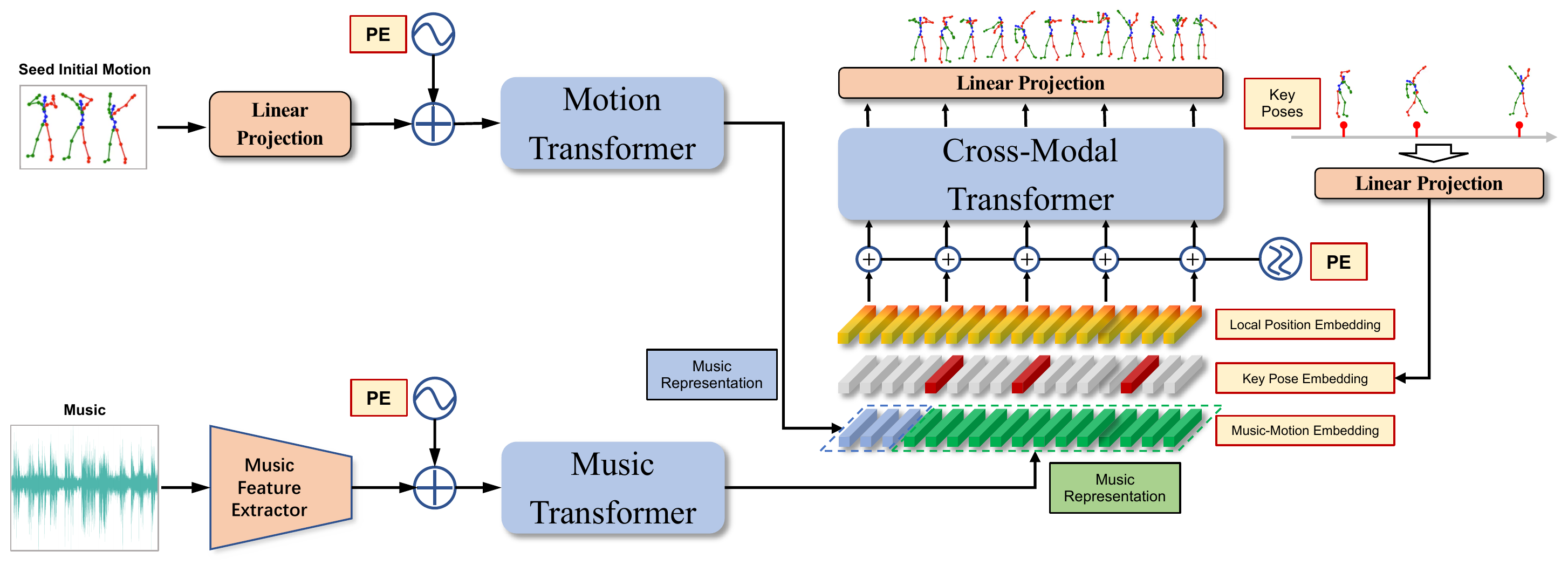}
  \caption{Overview of our dance synthesis framework. The system consists of two transformer encoders for motion and music embedding, 
  as well as a cross-modal transformer to learn the correspondence between the cross-modal data and generate the sequence of dance motion with key poses constraint. The positions of key poses expected to be appeared in the generated motion are encoded by a local position embedding mechanism,
  leading a position prior which makes the output poses at key frames have a better consistency with the input key poses.}
  \label{fig:framework}
  \vspace{-1em}
\end{figure*}

\section{Related Work}
\label{sec:related}

\subsection{Human Motion Synthesis}
\label{subsec:human_motion_synthesis}
Predicting future frames with initial pose or past motion has attracted researchers for a long time.
Early works mainly focus on statistical sequential models for motion sequence prediction.
Galata \etal~\cite{galata2001learning} make use of variable length Markov models for the efficient behaviours representation, and achieve good performance for long-term temporal motion prediction.
Brand \etal~\cite{brand2000style} approach the problem of stylistic motion synthesis by learning motion patterns from highly varied of motion sequences with style-specific Hidden Markove Models (HMMs).
Different from motion prediction with initial pose, Petrovich \etal~\cite{petrovich2021action} attempt to synthesize motion from scratch conditioned on action categories.
They design a Transformer-based architecture for encoding and decoding a sequence of human poses parameterized by SMPL model in conjunction with generative VAE training.

\subsection{Human Pose and Motion Modeling}
A recent trend in 3D human pose and shape estimation is to use deep neural network to learn a parametric human body representation.
The Skinned Multi-Person Linear Model (SMPL) is a realistic 3D model of the human body that is wildly used for various tasks, \eg, pose estimation~\cite{kolotouros2019learning}, motion synthesis~\cite{petrovich2021action}, action recognition~\cite{varol2021synthetic}, \etc.
To have a robust representation of human body, Graph neural network (GNN) is a good choice due to the intrinsic graph structure of human body.
For motion sequence modeling, deep recurrent neural networks (RNNs) and transformer-based networks show the powerful capability for sequence processing and achieve top performance in different tasks.

\subsection{Cross-Modal Sequence Generation}
Except for motion synthesis with action category conditioned, sequence generation task with cross-modal data is becoming a research trend, 
which involves vision, audio, text, \etc.
In natural language processing and computer vision, a subset of prior works attempt to translate text instructions to action motions for virtual agents~\cite{hatori2018interactively}.
Text-based motion synthesis targets on generating realistic motion sequence following the semantic meaning of the input text.
The cross-modal data are expected to be semantically aligned while generating motions.
Ahuja~\cite{ahuja2019language2pose} propose to learn a joint embedding space for pose and language with curriculum learning.
Semantic understanding from input text plays an important role in this task.
Ghosh \etal~\cite{ghosh2021synthesis} propose to use a hierarchical two-stream sequential model to explore a finer joint-level mapping between input sentences and pose sequences.
Similar to natural language, audio data is also used as input for motion generation.
Music-driven dance synthesis is a difficult task due to the complicated choreography.
To generate long sequences of realistic dance motion, Transformer-based architectures are utilized in many prior works~\cite{li2021ai,huang2020dance,li2021dancenet3d}.

\section{Approach}
\label{sec:approach}
In this section, we present our approach to generate dance driven by music with key pose constraint. 
The proposed framework involves two transformer encoders for motion and music representation, respectively.
To explore the correspondence between music and motion sequences, a cross-modal transformer is utilized to fuse such two representations and generate dance motions.

\subsection{Framework and Formulation}
\label{sec:formulation}
We briefly discuss the formulation of our task.
Here we have a piece of music represented as $\bm{X} = \left\{x_t \right\}_{t=1}^T$, 
and a seed sequence of motion represented as $\bm{Y} = \left\{y_t\right\}_{t=1}^{T^{'}}$, 
as well as a sequence a key poses $\hat{\bm{Y}} = \left\{\hat{y}_{t_i} \right\}_{i=1}^M$,
where $T \gg T^{'}$, $T$ and $T^{'}$ are the lengths of $\bm{X}$ and $\bm{Y}$, respectively. 
$M$ is the number of key poses in sequence $\hat{\bm{Y}}$ while $t_i$ is the timestamp corresponding to key pose $\hat{y}_{t_i}$, 
where $t_M \leqslant T$.
Our goal is to generate the motion sequence $\widetilde{\bm{Y}} = \left\{\widetilde{y}_t \right\}_{{t=T^{'}+1}}^T$ from time step $T^{'}+1$ to $T$, with the constraint that the key poses expected to appear in the generated motion $\widetilde{\bm{Y}}$ at corresponding time step,
which means the mean error, represented as 
$\mathcal{E}=\frac{1}{M} \sum\limits_{i=1}^{M} ||\widetilde{y}_{t_i} - \hat{y}_{t_i}||^2$,
between the input key poses and generated key poses 
should be as smaller as possible. 

\cref{fig:framework} illustrates the framework of our dance motion generation framework. 
There are multiple inputs to the model, including a piece of driven music, a segment of initial seed motion clip, and sequence of key poses with the corresponding positions.
Our dance generation framework consists of the following tiers of neural network.
\begin{enumerate}[1)]
\item \textbf{Transformer encoders for motion and music.} To generate dance motions, we need a representative embedding for music features. 
    we utilize transformer to encode music considering its powerful capability of sequence modeling.
    We use a two-stream transformer architecture to encode music and motion, respectively.
\item \textbf{Motion generation with cross-modal transformer.} The motion and music embedding from the transformer encoders are concatenated 
    along time dimension and sent to a cross-modal transformer to learn the correspondence between these two modalities.
    The output of the transformer, which involves choreography contained in the correspondence between motion and music embedding, is projected to pose space with a linear layer as the predicted motion.
\end{enumerate}

\subsection{Transformer Encoders for Music and Motion}
The driven music and initial seed motion are represented as $\bm{X} = \left\{x_t \right\}_{t=1}^T$ and $\bm{Y} = \left\{y_t\right\}_{t=1}^{T^{'}}$, respectively.
The input of music transformer contains a set of music features extracted from raw audio including 12-D chroma, 2-D downbeat, 1-D onset.
We concatenate all these features resulting in a 15-D music representation, \eg~$\bm{X}$.
Transformer encodes the sequence with fully attention among all time steps, which makes it wildly used in modeling sequence with long-term dependency.
The transformer consists of scaled dot-product attention units.
The scaled dot-product attention deals with three inputs, including queries, keys, and values, represented as $Q$, $K$, and $V$, respectively.
These inputs are firstly embedded into the same feature space of dimension $D$ by a linear layer.
Denote the function of linear layer as $\varphi(\cdot)$, the attention weights $W$ are calculated as follows, 
\begin{equation}
    W = \rm{Softmax}\left( \frac{\varphi_q(Q)\varphi_k(K)^T}{\sqrt{D}} \right),
\end{equation}
The output of scaled-dot production attention $O_A$ is as follows,
\begin{equation}
    O_A = W\times\varphi_v(V).
\end{equation}
In transformer, masking is a skillful mechanism for different task, which controls the attention relationship among the whole sequence.
For the task of sequence generation with auto-regression, the mask is designed as upper
triangular to enable casual attention.
The music features are first embedded in a high-dimension latent space via linear projection $\varphi_a(\cdot)$.
The queries, keys, and values of music transformer are all the same and set to be the output of linear layer.
We use $\mathcal{F}_{Tr}^{(A)}(\cdot)$ to represent music transformer \footnote{To avoid confusing with motion transformer and simplify the notation, we use the superscript ``A" corresponding to ``audio'' to represent music related module.}, 
then the output of music transformer encoder is denoted as follows,
\begin{equation}
\label{eq:emb_music}
    \bm{E}^{(A)} = \left\{ e_t^{(A)} \right\}_{t=1}^{T} = \mathcal{F}_{Tr}^{(A)} \left( \left\{\varphi_a(x_t)\right\}_{x=1}^{T} \right).
\end{equation}

The data of each frame in the sequence of dance motion consists of human body pose and the global translation vector.
The body pose is represented by 24 joints with 3-D rotation angle, and the 3-D translation vector indicates the global position of root joint.
We use the 6-D rotation representation proposed in~\cite{zhou2019continuity} considering its continuity.
Thus, the motion data in each frame is represented as a 147-D vector by concatenating the rotation vector and translation vector.
Given a sequence of initial dance motion clip, denoted as $\bm{Y} \in \mathbb{R}^{T^{'}\times 147}$, we first embed the input into a high-dimension feature space with a linear layer $\varphi_p(\cdot)$.
The the motion transformer followed by the linear layer models the temporal correlations of input embedding, generating a sequence of motion representations with the attention among the whole input sequence.
In a similar way with music transformer, 
we use $\mathcal{F}_{Tr}^{(M)}(\cdot)$ to represent motion transformer, then the output of motion transformer encoder is denoted as follows,
\begin{equation}
\label{eq:emb_motion}
    \bm{E}^{(M)} = \left\{ e_t^{(M)} \right\}_{t=1}^{T^{'}} = \mathcal{F}_{Tr}^{(M)} \left( \left\{\varphi_p(y_t)\right\}_{t=1}^{T^{'}} \right).
\end{equation}

\subsection{Motion Generation}
To learn the correspondence between music and motion,
the representations of music and motion are concatenated along time dimension as one of the inputs of cross-modal transformer.
The cross-modal transformer models multiple inputs including the joint concatenation of music and motion embedding, the key pose embedding, as well as the local position embedding where the key pose expected to be located in the output motion sequence.

\subsubsection{Key pose embedding.}
For the key poses, we only focus on the posture of human body while ignoring the global translation.
The trajectory which consisted of the global translation has potential effect on the overall feelings of the generated dance motion.
However, the translation of the isolated key pose is pointless, even has restrictions on generating dance motions.
In consideration of the aspects mentioned above, we choose to concentrate on the posture of human body with only joint rotations and discard the global translations.
We embed the key poses into a feature vector which has the same dimension with the music and motion representations calculated by \cref{eq:emb_music} and \cref{eq:emb_motion} via a linear projection layer.
The positions without key poses are padded with zeros.
The sequence of key pose embedding is represented as follows,
\begin{equation}
\label{eq:emb_keypose}
    \bm{E}^{(P)} = \left\{ e_t^{(P)} \right\}_{t=1}^T,
\end{equation}
where
\begin{equation}
e_t^{(P)} =\left\{
\begin{aligned}
    \varphi(\hat{y}_{t}) & , & t \in \left\{t_i ; 1\le i \le M \right\}, \\
    \bm{0} & , & otherwise,
\end{aligned}
\right.
\end{equation}
$\hat{y}_{t} \in \hat{\bm{Y}}$ is the key pose introduced in \cref{sec:formulation}.

\subsubsection{Local Positional Embedding.}
Unlike Recurrent Neural Networks (RNNs) that process the temporal sequence one by one with recurrence mechanism, 
Transformer ditches the ordinal temporal inputs in favor of full attention mechanism crossing the entire sequence.
Although such kind of designing can capture long dependencies, and accelerate the training and inference procedure due to the parallel computation, 
there exists deficiency of the model since the temporal-order information is omitted.
To make the model has sense of order information, an alternative solution is incorporate the position representation into the input sequence.

\begin{figure}[t!]
  \centering
  \includegraphics[width=1.\linewidth]{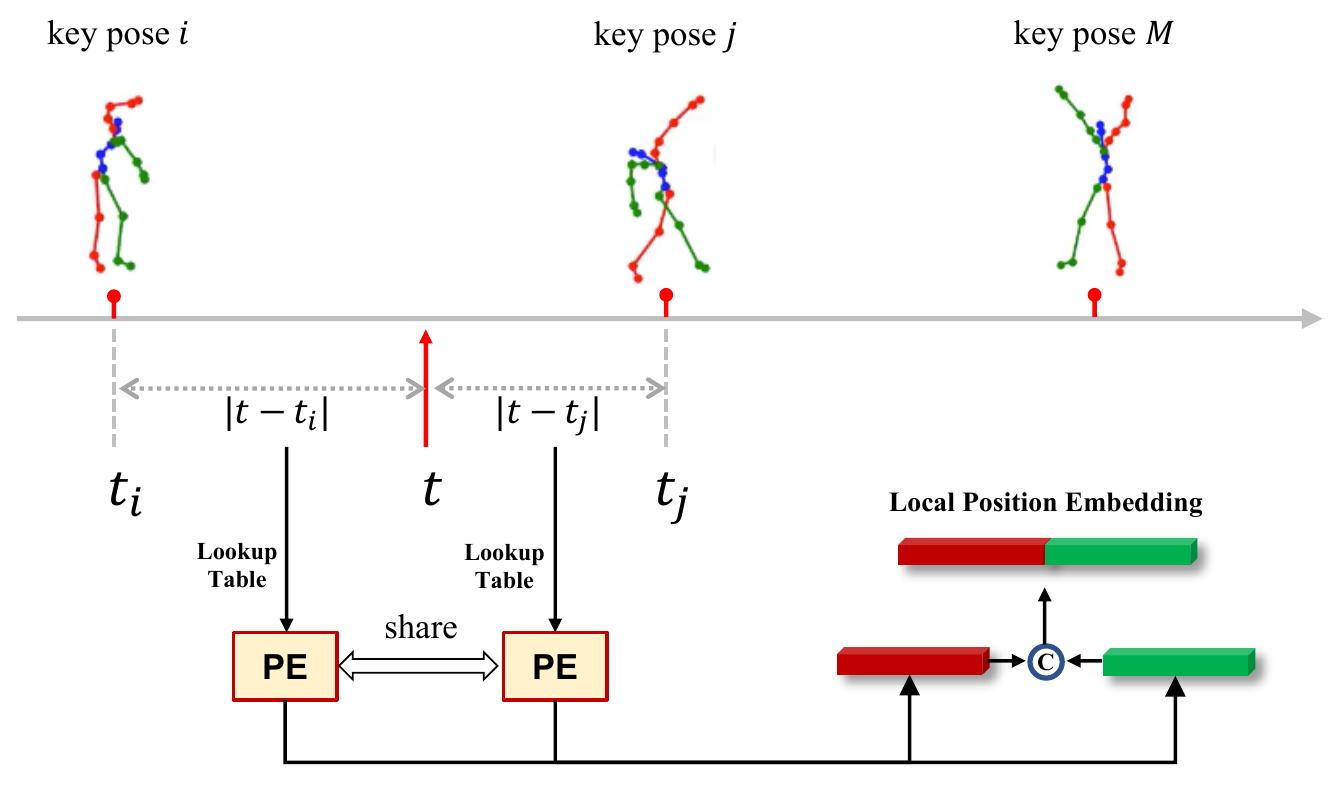}
  \caption{Illustration of local positional embedding. The relative positions of adjacent key poses are encoded to provide the distribution of key poses.}
  \label{fig:lpe}
\end{figure}

In the task of motion generation with key poses, the output pose of each frame is modeled by the whole sequence of input poses.
However, the key poses with different positions have different contributions to the output pose.
In general, the significance for the output pose of current frame depends on the distance between current pose and key pose.
The key poses, which are closer to output pose, contribute more to the prediction.
Based on such consideration, we propose a local position embedding module to incorporate the position information of the adjacent key poses in both sides.
The local position embedding mechanism is illustrated in~\cref{fig:lpe}.
We briefly review the ordinary positional encoding in~\cite{vaswani2017attention}.
The position information is encoded with a positional embedding matrix $PE \in \mathbb{R}^{N\times d_{model}}$, 
where $N$ is the potential maximum length of sequence, $d_{model}$ is the input dimension of the Transformer.
The element on $pos^{th}$ row and $i^{th}$ column in $PE$ is represented as follows,
\begin{equation}
\label{eq:pe}
    \begin{split}
        PE(pos, 2i) &= \sin(pos / 10000^{2i/d_{model}}), \\
        PE(pos, 2i+1) &= \cos(pos / 10000^{2i/d_{model}}).
    \end{split}
\end{equation}

The local positional embedding is derived with $PE$.
For the current input at time step $t$, we denote the left and right adjacent key poses as $\widetilde{y}_i$ and $\widetilde{y}_j$ at the temporal position of $t_i$ and $t_j$, respectively.
The local positional embedding consists of two parts corresponding to relative positions of left and right adjacent key poses.
We get the ordinary positional embedding ${pe}_l$ of the key pose $\widetilde{y}_i$ in left side by indexing the positional embedding matrix $PE$ with relative temporal distance $|t-t_i|$, represented as 
\begin{equation}
    {pe}_l = {PE}(|t-t_i|, *).
\end{equation}
Similarly, ${pe}_r$ of the key pose $\widetilde{y}_j$ in right side is as follows,
\begin{equation}
    {pe}_r = {PE}(|t-t_j|, *).
\end{equation}
The local relative positional embedding $pe^{(L)}_n$ for time step $t$ is the concatenation of ${pe}_l$ and ${pe}_r$, represented as follows,
\begin{equation}
    {pe}^{(L)}_n = [{pe}_l, {pe}_r],
\end{equation}
where $[\cdot, \cdot]$ the concatenation operation.

\begin{figure}[t!]
  \centering
  \includegraphics[width=1.0\linewidth]{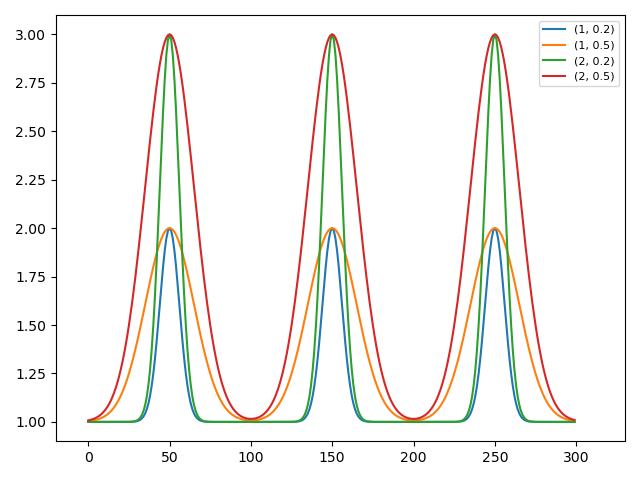}
  \caption{An example for the weight curve $\omega_t$ with different paired parameters $(\lambda, \sigma)$.}
  \label{omega}
\end{figure}

\subsubsection{Cross-Modal Transformer.}
We use a cross-modal Transformer to learn the correlation between initial motion embedding and music embedding.
The motion representation $\bm{E}^{(M)}$ and music representation $\bm{E}^{(A)}$ are concatenated along time axis as one of the cross-modal Transformer inputs.
Other information cues that expected to be modeled by the cross-modal Transformer include key pose embedding and local positional embedding.
Thus, the overall input of the cross-modal Transformer is represented as the summation of local positional embedding, key pose embedding, as well as the music-motion joint embedding. Denote the overall input as $\bm{E}^{(C)}_{in}$, which is derived as follows,
\begin{equation}
    \bm{E}^{(C)}_{in} = [\bm{E}^{(M)}, \bm{E}^{(A)}] + \bm{E}^{(P)} + \bm{PE}^{(L)} + \bm{PE},
\end{equation}
where $\bm{E}^{(M)}$, $\bm{E}^{(A)}$, $\bm{E}^{(P)}$ corresponds to embeddings of motion, music, and key poses, respectively.
$\bm{PE}^{(L)}$ is the local positional embedding introduced in the previous section, $\bm{PE}$ is the ordinary position encoding calculated in \cref{eq:pe}.
We denote the mapping function of the cross-modal Transformer as $\mathcal{F}_{Tr}^{(C)}(\cdot)$.
By passing the overall input embedding $\bm{E}^{(C)}_{in}$ through  $\mathcal{F}_{Tr}^{(C)}(\cdot)$, we obtain the output sequence 
$\bm{E}^{(C)}_{out} = \mathcal{F}_{Tr}^{(C)} (\bm{E}^{(C)}_{in})$.
We use a linear transformation layer $\varphi_o(\cdot)$ to transpose the output sequence into the final predicted motions which contain a sequence of $T$ poses and global translation vectors, the final predictions are as follows,
\begin{equation}
    \widetilde{\bm{Y}} = \varphi_o(\bm{E}^{(C)}_{out}).
\end{equation}

\subsection{Training}
\label{sec:training}
To optimize the network, we design a weighted reconstruction loss on motions.
We define the consistency error $\mathcal{E}_c$ between the key poses and the corresponding generated poses at the same time steps.
Given a sequence of key poses $\hat{\bm{Y}} = \left\{\hat{y}_{t_i} \right\}_{i=1}^M$ and the predicted motions $\widetilde{\bm{Y}} = \left\{\widetilde{y}_t \right\}_{t=1}^T$, the consistency error $\mathcal{E}_c$ is calculated by mean squared error (MSE) as follows,
\begin{equation}
    \mathcal{E}_c=\frac{1}{M} \sum\limits_{i=1}^{M} ||\widetilde{y}_{t_i} - \hat{y}_{t_i}||^2,
    \label{eq:error_consistency}
\end{equation}
where $M$ is the number of key poses, $t_i$ corresponds to the time step of $i^{th}$ key pose.
As illustrated in the previous section, our goal is to generate the motion sequence with the constraint that the key poses expected to appear in the generated motion at corresponding time step, which means $\mathcal{E}_c$ is supposed to be as smaller as possible.
To this end, the error at the key pose position is enlarged by multiply a large factor.
The weighted reconstruction loss between the predicted motions $\widetilde{\bm{Y}} = \left\{\widetilde{y}_t \right\}_{{t=1}}^T$ and ground-truth sequence $\overline{\bm{Y}}_{gt} = \left\{\overline{y}_t \right\}_{{t=1}}^T$.
The weighted reconstruction loss $\mathcal{L}$ is calculated as follows,
\begin{equation}
    \mathcal{L} = \frac{1}{T} \sum\limits_{t=1}^{T} \omega(t)\cdot ||\overline{y}_{i} - \hat{y}_{i}||^2,
    \label{eq:loss_rec}
\end{equation}
where
\begin{equation}
    \omega(t) = \sum\limits_{i=1}^{M} 1 + \lambda e^{-\frac{(t-t_i)^2}{2\sigma^2}}.
\end{equation}
$\lambda$ is a tunable hyper-parameter which balances the smoothness of the whole dance motion sequence and consistency with the key poses.
$\sigma$ controls the range that is affected by the key poses.
An example for a fixed distribution of key poses with different paired parameters denoted as $(\lambda, \sigma)$ is shown in~\cref{omega}.
From this figure, we notice that the maximum weight appears at the key frame where the key pose located in.
The weight decreases when it leaves far from the key pose.
A small $\sigma$ means a sharp decline of the weight for the calculation of reconstruction loss defined in \cref{eq:loss_rec}.
Obvious, there is a trade-off between the smoothness of the whole generated dance motions and the consistency with the key poses.
With the increase of $\lambda$, the model learns to generate dance motions with a smaller consistency error $\mathcal{E}_c$, which makes the output poses at the key frame much more closer to the input key poses.

\section{Experiments}
\subsection{Dataset and Implementation Details}
We conduct our experiments on \emph{AIST++}~\cite{li2021ai}, a large-scale 3D human dance motion dataset consists of 3D motion paired with music.
A variety of cross-modal analysis and synthesis tasks are utilized with \emph{AIST++}, including dance generation conditioned on music, human motion prediction, \etc.
We split the dataset for training and evaluation following the setup in ~\cite{li2021ai}.

We use Adam optimizer~\cite{kingma2014adam} with a initial learning rate of 1e-4 and 20 batch size.
The learning rate drops to 1e-5 and 1e-6 after 100k and 250k iterations, respectively.
We train two models with different number of parameters.
For the large model, the cross transformer has 12 layers with 10 attention heads and 800 hidden size for each layer.
For the light model, the number of layers and hidden size are reduced, which is set to be 8 layers with 4 attention heads and 256 hidden size for each layer.
The motion transformer and music transformer in both large model and light model have the same layers which is set to be 4.
The whole framework is implemented on PyTorch and experiments are performed on NVIDIA Tesla V100 GPU.

\subsection{Experimental Results}
\subsubsection{Evaluation metric.}
We utilize two different evaluation metrics for dance generation on key pose consistency and smoothness, respectively.
On one hand, the synthesized dance motion is controlled by the given key poses.
As a result, it's significant to keep the generated poses at key frames as similar as possible with the given key poses.
Hence, we use the mean squared error (MSE) between the generated key poses and the input key poses, formulated in \cref{eq:error_consistency}.
On the other hand, smoothness is one of the most important evaluation metrics in motion synthesis task.
For the smoothness evaluation, we utilize the coefficient of variation $S_{cv}$, defined as the ratio of the standard deviation to the mean, for the temporal difference of the generated pose sequence, 
which is formulated as follows,
\begin{equation}
    S_{cv} = \frac{sd(D_f(\widetilde{\bm{Y}}))}{|mean(D_f(\widetilde{\bm{Y}}))|},
\end{equation}
where $sd(\cdot)$ and $mean(\cdot)$ correspond to standard deviation and mean value of the sequence, respectively.
$D_f(\widetilde{\bm{Y}}) = \left\{\widetilde{y}_t \right\}_{{t=1}}^T - \left\{\widetilde{y}_t \right\}_{{t=0}}^{T-1}$ 
means the temporal difference of the generated motion sequence.
A smaller smoothness score $S_{cv}$ means the generated dance tend to be more smooth.
In addition, we use beat hit rate to evaluate whether the dance motion beats and musical beats are aligned to each other.
The beat hit rate is defined as $N_{d\&m} / N_m$, 
where $N_{d\&m}$ is the number of dance motion beats that are aligned with musical beats, and $N_m$ is the number of musical beats.
The aligned dance motion beat is defined whether the musical beat occurs in its adjacent temporal interval $[-\delta/f_d, \delta/f_d]$,
where $f_d$ is the $FPS$ of dancing video, $\delta$ is a hyper-parameter to control the range of the temporal interval.

\subsubsection{Study on different hyper-parameters.}
\begin{figure}[]
  \centering
  \includegraphics[width=1\linewidth]{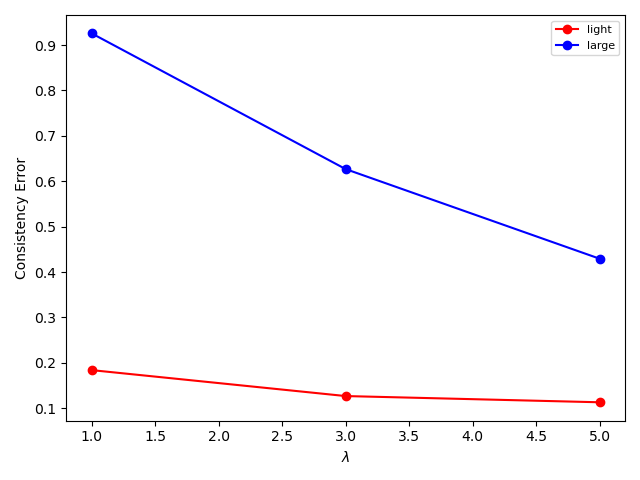} 
  \caption{Consistency error with different $\lambda$ on large/light model. With the increase of $\lambda$, the consistency error of the key poses gradually decreases.}
  \label{fig:consistency_lambda}
\end{figure}

Our model is optimized with weighted reconstruction loss defined in~\cref{eq:loss_rec}.
We have two hyper-parameters in this equation.
The results with different combination of $\lambda=\{0, 1, 3, 5\}$ and $\sigma=\{0.05, 0.1, 0.2\}$ are given in~\cref{tab:params_light}-\cref{tab:cov1_params_large}.
From the results in~\cref{tab:params_light} and~\cref{tab:params_large}, we notice that the hyper-parameter $\sigma$ has less effect on the consistency error compared with $\lambda$.
With the increase of $\lambda$, the consistency error of the key poses gradually decreases, also as shown in~\cref{fig:consistency_lambda}.
This is reasonable, since larger $\lambda$ enforce the network to pay more attention to the error at the key frame.

\begin{table}[ht!]
\begin{center}
\caption{Consistency error with different hyper-parameter $\sigma$ and $\lambda$ on light model.}
\label{tab:params_light}
\tabcolsep=5.5pt
\begin{tabular}{lc|ccccc}
\hline
\multicolumn{2}{c}{\multirow{2}{*}{}} & \multicolumn{5}{|c}{$\lambda$} \\
\multicolumn{2}{c|}{}                  & 0    & 1    &  3   &    5   & Mean \\ \hline
\multirow{4}{*}{$\sigma$} & 0.05  & 0.2521     &  0.1867     & 0.1130   &  0.0839  &  0.1589    \\
   & 0.1                          & 0.2421     &  0.1869     & 0.1224   &  0.1072  &  0.1647    \\
   & 0.2                          & 0.2504     &  0.1785     & 0.1454   &  0.1485  &  0.1807    \\ 
   & Mean                         & 0.2482     &  0.1840     & 0.1269   &  0.1132  & -          \\ \hline
\end{tabular}
\end{center}
\end{table}

\begin{table}[ht!]
\begin{center}
\caption{Smoothness score $S_{cv}$ with different hyper-parameter $\sigma$ and $\lambda$ on light model.}
\label{tab:cov1_params_light}
\tabcolsep=5.3pt
\begin{tabular}{lc|ccccc}
\hline
\multicolumn{2}{c}{\multirow{2}{*}{}} & \multicolumn{5}{|c}{$\lambda$} \\
\multicolumn{2}{c|}{}                  & 0    & 1    &  3   &    5   & Mean \\ \hline
\multirow{4}{*}{$\sigma$}  & 0.05  & 151.25     &  166.67     & 171.22   &  114.53  &  150.92    \\
                           & 0.1   & 262.48     &  119.45     & 216.44   &  249.41  &  211.95    \\
                           & 0.2   & 340.91     &  172.65     & 143.84   &  320.01  &  244.35  \\ 
                           & Mean  & 251.55     &  152.92     & 177.17   &  227.98  & -  \\ \hline
\end{tabular}
\end{center}
\end{table}

\begin{table}[ht!]
\begin{center}
\caption{Consistency error with different hyper-parameter $\sigma$ and $\lambda$ on large model.}
\label{tab:params_large}
\tabcolsep=5.5pt
\begin{tabular}{lc|ccccc}
\hline
\multicolumn{2}{c}{\multirow{2}{*}{}} & \multicolumn{5}{|c}{$\lambda$} \\
\multicolumn{2}{c|}{}                  & 0    & 1    &  3   &    5   & Mean \\ \hline
\multirow{4}{*}{$\sigma$} & 0.05  & 1.7807     &  0.9604     & 0.4113   &  0.2544  &  0.8517    \\
   & 0.1                          & 1.7560     &  0.8441     & 0.5656   &  0.3605  &  0.8816       \\
   & 0.2                          & 1.8335     &  0.9721     & 0.9032   &  0.7247  &  1.1084  \\ 
   & Mean                         & 1.7901     &  0.9255     & 0.6267   &  0.4294  & -  \\ \hline
\end{tabular}
\end{center}
\end{table}

\begin{table}[ht!]
\begin{center}
\caption{Smoothness score $S_{cv}$ with different hyper-parameter $\sigma$ and $\lambda$ on large model.}
\label{tab:cov1_params_large}
\tabcolsep=5.2pt
\begin{tabular}{lc|ccccc}
\hline
\multicolumn{2}{c}{\multirow{2}{*}{}} & \multicolumn{5}{|c}{$\lambda$} \\
\multicolumn{2}{c|}{}                  & 0    & 1    &  3   &    5   & Mean \\ \hline
\multirow{4}{*}{$\sigma$}  & 0.05  & 429.44    &  280.76     & 301.89   &  378.83  &  347.74    \\
                           & 0.1   & 297.52    &  149.51     & 162.49   &  191.84  &  200.34    \\
                           & 0.2   & 220.45    &  189.05     & 185.77   &  365.48  &  240.19  \\ 
                           & Mean  & 315.80    &  206.44     & 216.72   &  312.05  & -  \\ \hline
\end{tabular}
\end{center}
\end{table}

\begin{figure}[]
  \centering
  \includegraphics[width=1\linewidth]{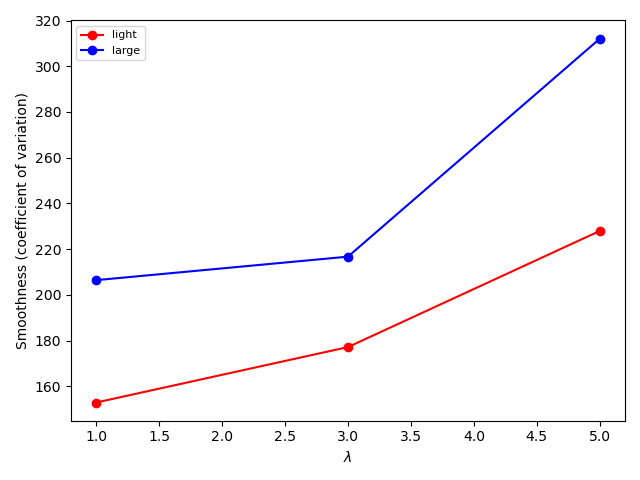} 
  \caption{Smoothness score with different $\lambda$ on large/light model. With the increase of $\lambda$, the smoothness score gradually increase.}
  \label{fig:smooth_lambda}
\end{figure}

As for the smoothness scores in~\cref{tab:cov1_params_light} and~\cref{tab:cov1_params_large}, 
we can see that the smoothness score moves in the opposite direction to the consistency error, as shown in~\cref{fig:smooth_lambda}.
With the increase of $\lambda$, the smoothness score gradually increase as well,
which means a large value of $\lambda$ could have negative effects on the smoothness of the synthesized dancing videos.
The experimental results and the above analysis indicate that there is a trade-off between the consistency error with the key poses and the smoothness score of the synthesized dance motion, mentioned in~\cref{sec:training} as well.
Although a large value of $\lambda$ enforce the generated key poses to have a high similarity with the given anchor poses,
it does have negative effects on the smoothness of the whole synthesized dancing video.
We can adjust the parameter $\lambda$ to generated dance motions determined by our preferences for consistency or smoothness.

\begin{table}[ht!]
\begin{center}
\caption{Beat hit rate (\%) with different $\delta$ (fixed $\lambda=3$ and $\sigma=0.1$, the higher the better.)}
\label{tab:beat_hitrate}
\tabcolsep=7pt
\begin{tabular}{lc|ccccc}
\hline
\multicolumn{2}{c}{\multirow{2}{*}{}} & \multicolumn{5}{|c}{$\delta$} \\
\multicolumn{2}{c|}{Dance}                  & 1    & 2    &  3   &    4   & 5 \\ \hline
                           & Real  & 10.4     &  25.7    & 36.3   &  42.6  &  48.7  \\
                           & Synthesized      &  7.7     &  19.5    & 25.9   &  34.2  &  37.6  \\ \hline
\end{tabular}
\end{center}
\end{table}

\begin{figure}[t]
  \centering
  \includegraphics[width=1.0\linewidth]{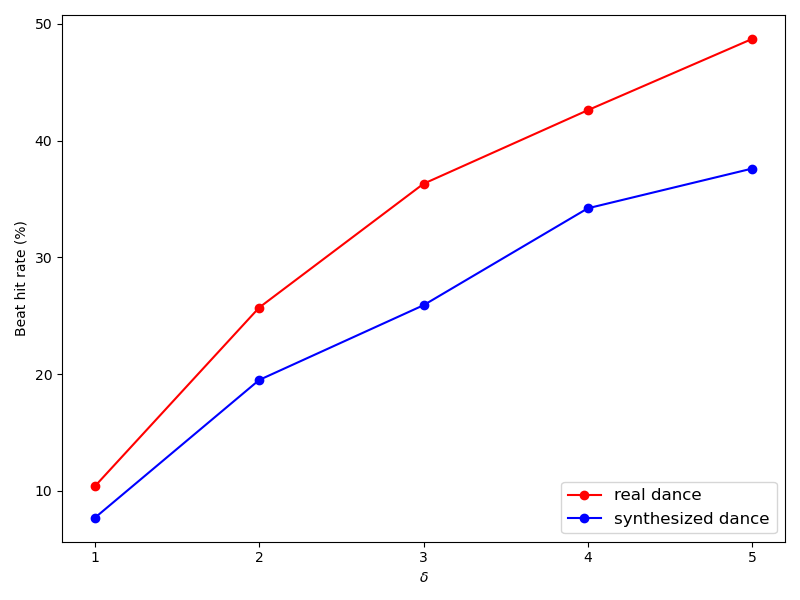}
  \caption{Beats hit rate with different $\delta$ on real dance and synthesized dance.}
  \label{fig:beathitrate}
\end{figure}

\begin{figure*}[ht!]
  \centering
  \includegraphics[width=1\linewidth]{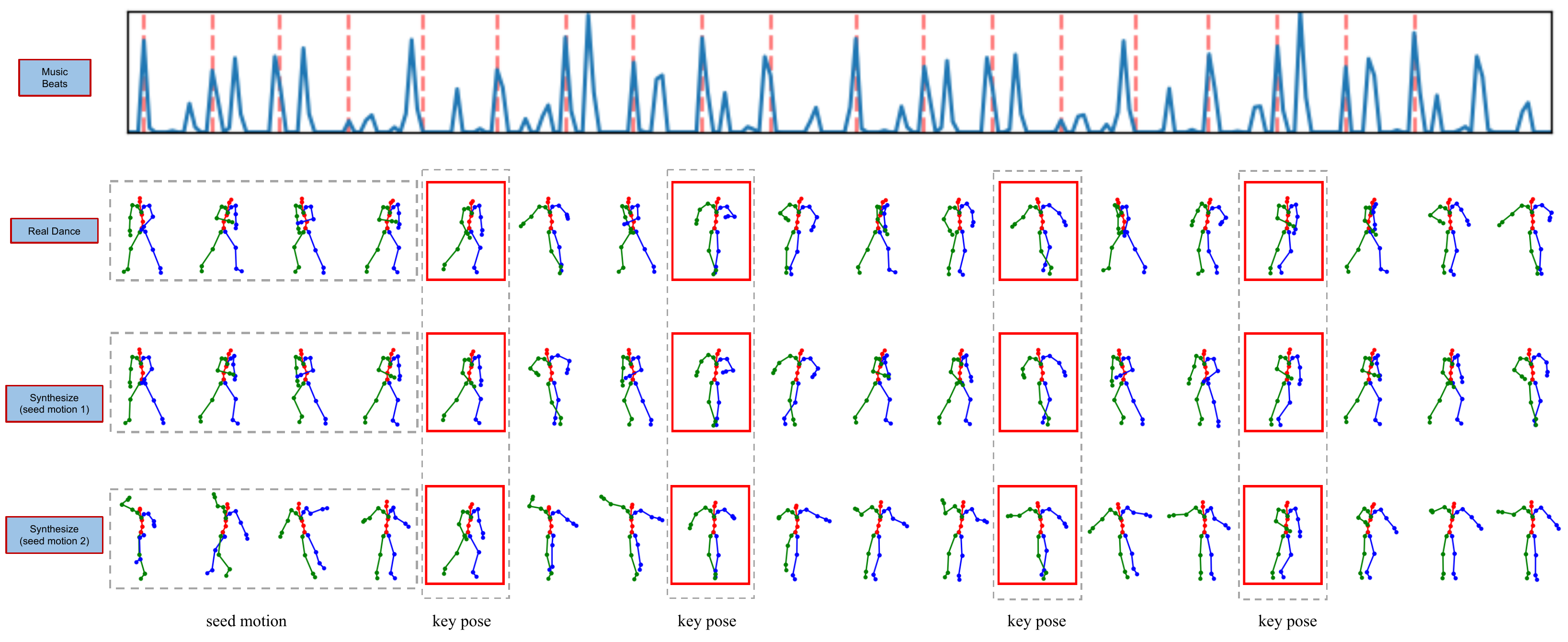}
  \caption{Dance motions generation results controlled by the key poses with different seed motions. The first row is the input music, the second row is the real dance motion. The following two rows show the generated dance motions with different seed motions. The key poses for both real dance motion and synthesized dance motion are marked with red bounding boxes. The generated poses at key frame have the same appearance with the input anchor key poses.}
  \label{fig:vis}
\end{figure*}

\subsubsection{Beat hit rate.}
The results of beat hit rate with different hyper-parameter $\delta$ are shown in~\cref{tab:beat_hitrate} and~\cref{fig:beathitrate}.
In our experiments, the video $FPS$ $f_d$ is set to be 30, and temporal range hyper-parameter $\lambda$ ranges from 1 to 5.
A small value of $\lambda$ indicates a strict alignment between dance motion beats and musical beats.
The beat hit rate of the synthesized dance motion gets close to the beat hit rate of the real dance motion,
which means our algorithm perform well on beat alignment of generated dance motion and input music.

\begin{figure}[t!]
  \centering
  \includegraphics[height=0.7\linewidth,width=1\linewidth]{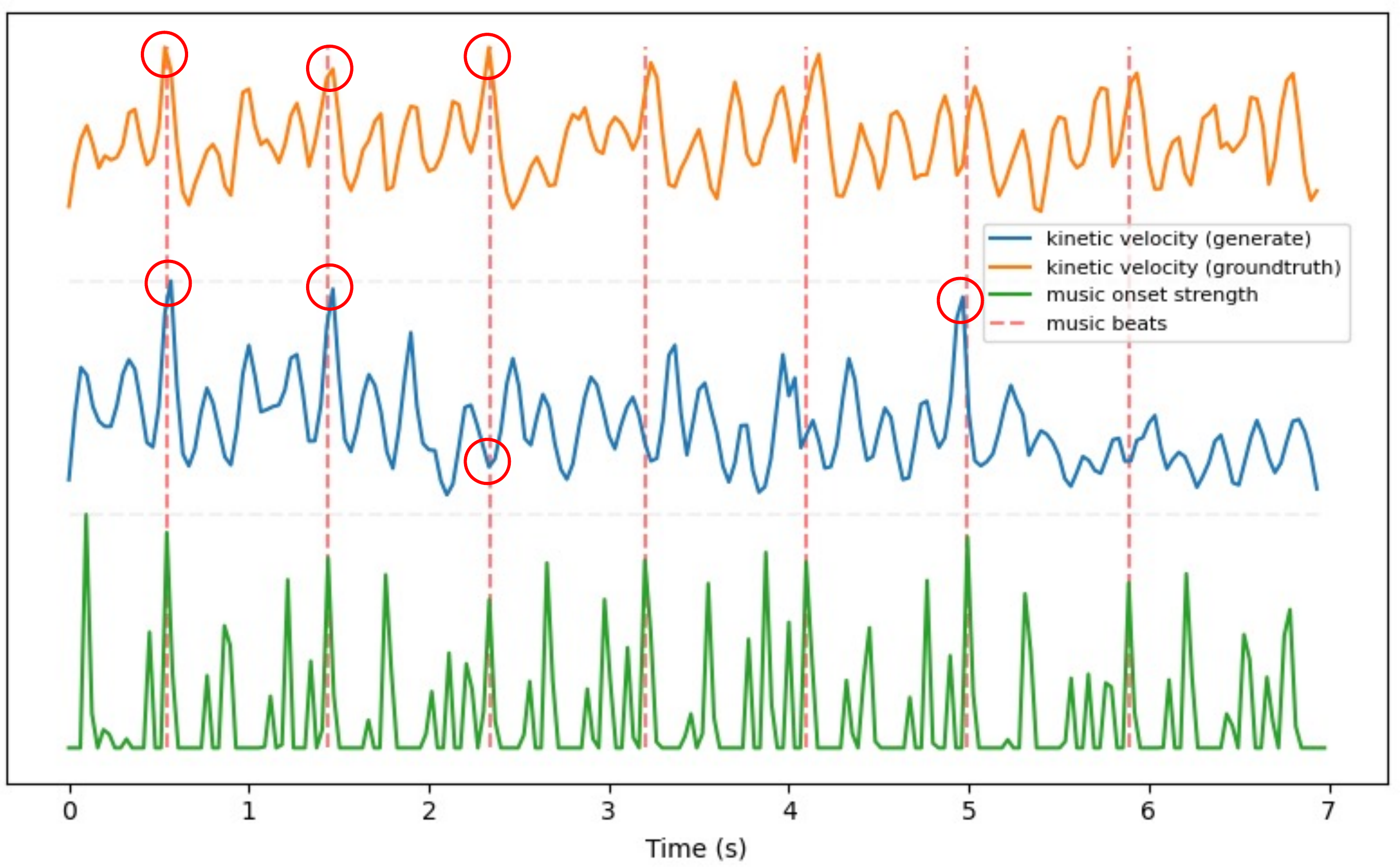}
  \caption{Beats alignment visualization. The curves from top to bottom are kinetic velocity of real dance (orange curve), kinetic velocity of synthesized dance (blue curve), music onset strength (green curve), respectively. The dance motion beats (red circle) are aligned with music beats (red dotted line).}
  \label{fig:beatsalign}
\end{figure}

\subsubsection{Visualization.} 
We visualize the synthesized dance motion sequences with different seed motions and controllable key poses constraints, shown in~\cref{fig:vis}.
We synthesize two dance motion sequences with different seed motions driven by the same music and a sequence of anchor key poses.
From the visualization result, we can see that the synthesized poses at key frame have the same appearance with the input anchor key poses, which keeps a good consistency, while the smoothness is guaranteed as well.
The diversity of the generated dance motions are achieved by the variations of the seed motions.
Besides, we visualize the alignment between dance motion beats and musical beats as well, shown in~\cref{fig:beatsalign}.
The dance motion beat is defined as the local extrema of the kinetic velocity. 
We mark the dance motion beats with red circles.
As we can see, the dance motion beats are aligned with music beats (orange dotted line).

\section{Conclusion}
In this paper, we propose a novel framework for dance motion synthesis based on music and key pose constraint. 
We target on synthesizing high-quality dance motion driven by music as well as customized poses performed by users.
Our model consists of single-modal transformer encoders for music and motion representations,
and a cross-modal transformer decoder for dance motions generation. 
We use a local positional embedding to encode the relative position information of adjacent key poses.
With this mechanism the cross-modal transformer are sensitive to key poses and the corresponding positions.
Our dance synthesis model achieves satisfactory performance both on quantitative and qualitative evaluations with extensive experiments, which demonstrates the effectiveness of our proposed method.

{\small
\bibliographystyle{ieee_fullname}
\bibliography{egbib}
}

\end{document}